\documentclass[11pt,a4paper]{article}

\usepackage[T1]{fontenc}
\usepackage[utf8]{inputenc}
\usepackage{lmodern}
\usepackage{microtype}
\usepackage{amsmath,amssymb,amsfonts,mathtools}
\usepackage{graphicx}
\usepackage{booktabs}
\usepackage[numbers,sort&compress]{natbib}
\usepackage{hyperref}
\usepackage[nameinlink,noabbrev]{cleveref}
\usepackage[margin=2.8cm]{geometry}
\usepackage{orcidlink}

\newcommand{\vect}[1]{\mathbf{#1}}
\newcommand{\set}[1]{\mathcal{#1}}

\hypersetup{
    colorlinks=true,
    linkcolor=blue,
    citecolor=blue,
    urlcolor=blue
}

\title{Structured Relevance Determination and Channel-Additive Gaussian Processes for Image Data}

\author{
	Harald Ro{\ss}els\,\orcidlink{0009-0006-8121-2639}\\
	Independent Researcher\\
	\texttt{harald.rossels@gmail.com}
	\and
	Ludovico Scarton\,\orcidlink{0000-0003-0174-2543}\\
	Bonn-Rhein-Sieg University of Applied Sciences\\
	\texttt{ludovico.scarton@gmail.com}
}
\date{Preprint --- October 2026}

\usepackage[acronym]{glossaries}
\newacronym[
    plural=GPs,
    longplural={Gaussian processes}
]{gp}{GP}{Gaussian process}
\newacronym{rbf}{RBF}{radial basis function}
\newacronym{ard}{ARD}{automatic relevance determination}
\newacronym{acrd}{ACRD}{automatic channel relevance determination}
\newacronym{aprd}{APRD}{automatic patch relevance determination}
\newacronym{elbo}{ELBO}{evidence lower bound}
\newacronym{nlpp}{NLPP}{negative log predictive likelihood}

\begin{document}

\maketitle

\begin{abstract}
We introduce radial basis function (RBF) kernels for Gaussian processes that use structured hyperparameter sharing for image-valued inputs. 
For high-dimensional inputs such as images, a fully anisotropic RBF kernel assigns a separate lengthscale to each input dimension, which can introduce many hyperparameters and may lead to overfitting and poorly calibrated predictive uncertainty when data are limited. 
We propose structured relevance determination by tying lengthscales across groups of adjacent pixels and within each colour channel. 
This yields patch-wise and channel-wise constrained anisotropic RBF parametrisations that reduce the number of independently learned lengthscales. 
We further introduce channel-additive Gaussian processes for multi-channel images and combine them with structured relevance determination, resulting in a parameter-efficient decomposition of the latent function into components defined on channel subsets, enabling some non-local generalisation. 
The proposed kernels are evaluated on MNIST and a compact binary aircraft-detection dataset of low-resolution RGB satellite image chips. 
We also show how channel-wise structured relevance determination can be incorporated into convolutional Gaussian processes.
We observe improved predictive performance compared to isotropic and fully anisotropic RBF baselines.
\end{abstract}

\section{Introduction}\label{sec:introduction}

\glspl{gp} \citep{RasmussenWilliams2006} are a non-parametric Bayesian method for modelling distributions over functions and provide uncertainty estimates.
Their behaviour is largely determined by the kernel.
The \gls{rbf} kernel, also called squared exponential kernel, is a popular choice because it encodes a strong smoothness assumption and is widely used.
This local notion of similarity can become problematic for high-dimensional data when no additional structure is encoded in the kernel.
Bengio et al. \cite{Bengio2006CurseKernels} argue that methods relying only on smoothness and local kernels can be sensitive to highly variable target functions, which motivates kernel designs that encode additional non-local or problem-specific structure.
Additive \glspl{gp} \cite{Duvenaud2011AdditiveGPs} are one example of a kernel construction that can exploit lower-dimensional structure and enable non-local generalisation.
Convolutional \glspl{gp} \cite{VanDerWilk2017ConvGP} encode image-specific structure by constructing kernels from image patches, analogous to the use of convolutional structure in neural networks.
If one nevertheless wants to retain the \gls{rbf} kernel as a flexible prior over functions, an anisotropic parametrisation is one natural way to adapt it to high-dimensional inputs.
Since each dimension of the inputs to an anisotropic \gls{rbf} kernel is scaled by a dedicated lengthscale, this set of hyperparameters governs how quickly the kernel function decays in each direction and can make dimensions effectively irrelevant when the corresponding lengthscale becomes sufficiently large.
This is commonly referred to as \gls{ard}.
In the anisotropic case the number of lengthscales is equal to the dimensionality of the data points.
Learning a separate lengthscale for each input dimension can require substantially more data and may lead to overfitting and poorly calibrated predictive uncertainty when data are limited.
For many applications, such as settings involving decision-making, reliable uncertainty estimation can be more important than marginal improvements in point prediction accuracy.

Full \gls{ard} treats the input dimensions as unrelated coordinates. 
This is often unnecessarily general for structured high-dimensional inputs. 
In many applications, the coordinates of an input vector inherit relationships from an underlying domain, such as spatial neighbourhoods, sensor groups, physical regions, or different measurement channels. 
Such prior knowledge can be used to impose a sharing structure on kernel hyperparameters.
Image-valued inputs are a natural instance of such structured data. 
Pixels are arranged on a spatial grid, neighbouring pixels form local regions, and colour or spectral channels represent distinct components of the input. 
In a classification setting, connected clusters of pixels whose associated lengthscales are relatively small can be interpreted as regions where variations are of high importance for the discrimination between classes.
Alternatively, they can be seen as regions where important features, such as characteristic shapes, line segments, or colour patterns, are present.
Shared features across images of the same class usually still exhibit some variation with respect to position, i.e. translations.
In many image classification settings, evidence for a class is more likely to arise from spatially coherent patterns than from isolated single-pixel variations.
Consequently, an isolated pixel with a very small lengthscale while neighbouring pixels have much larger lengthscales may indicate fitting to pixel-level noise.

In this work, we introduce structured hyperparameter sharing for image-valued inputs. 
We first introduce \gls{aprd}, where lengthscales are shared across spatial patches, and \gls{acrd}, where lengthscales are shared across image channels. 
The resulting kernels can be interpreted as constrained \gls{ard} parametrisations of the \gls{rbf} kernel that reduce the effective number of hyperparameters. 
We further introduce channel-additive \glspl{gp} that decompose the latent function into components defined on individual channels and channel combinations.
This provides a form of non-local generalisation by allowing separate covariance contributions from channel subsets rather than relying only on a single distance over the full input.
Combining this additive construction with \gls{aprd} yields a parameter-efficient model that incorporates both channel-subset structure and shared spatial relevance.

\section{Background}\label{sec:background}

A \gls{gp} (see \cite{RasmussenWilliams2006} for an in-depth review)
\begin{equation}\label{eq:gp}
	f\left(\vect{x}\right) \sim \mathcal{GP}\left(m(\vect{x}), k\left(\vect{x}, \vect{x}^\prime\right)\right)
\end{equation}
is fully specified by its mean function $m$ and its covariance kernel function $k$.
In the following we assume centred \glspl{gp}, i.e. $m(\vect{x})=0$.

An anisotropic \gls{rbf} kernel is defined as
\begin{equation}\label{eq:rbfkernel}
	k(\vect{x}, \vect{x}^\prime) = \sigma^2\exp\left(-{\sum_{i=1}^{D}} \frac{\left(x_i - {x_i}^\prime\right)^2}{2{\ell_i}^2}\right)
\end{equation}
where each input dimension has its own lengthscale $\ell_i$.
The isotropic \gls{rbf} kernel is the special case where $\ell_i=\ell_j$ for all $i, j$.

Let $\left\{k_m\right\}_{m=1}^{M}$ be kernel functions defined over different spaces $\mathcal{X}_1, \dots, \mathcal{X}_M$, then
\begin{align}
	k\left(\vect{x}, \vect{x}^\prime\right) &= \sum_{m=1}^{M} k_m\left(\vect{x}_m, {\vect{x}^\prime}_m \right), \label{eq:sumkernel} \\
	k\left(\vect{x}, \vect{x}^\prime\right)  &= \prod_{m=1}^{M} k_m\left(\vect{x}_m, {\vect{x}^\prime}_m \right) \label{eq:productkernel}
\end{align}
are also kernel functions defined over the product space $\mathcal{X} = \mathcal{X}_1 \times \dots\times \mathcal{X}_M$.

The sum of independent \glspl{gp} is also a \gls{gp}.
Let $f_m\left(\vect{x}\right) \sim \mathcal{GP}\left(0,k_m\left(\vect{x}, {\vect{x}^\prime}\right)\right)$ with $m=1,\ldots,M$ be independent \glspl{gp}, then
\begin{equation}\label{eq:sumdecomposition}
	f\left(\vect{x}\right) = \sum_{m=1}^{M} f_m\left(\vect{x}\right) \Longrightarrow f\left(\vect{x}\right) \sim \mathcal{GP}\left(0, \sum_{m=1}^{M} k_m\left(\vect{x}, {\vect{x}^\prime} \right)\right).
\end{equation}

Sum and product compositions are common building blocks for designing \gls{gp} models with richer covariance functions.
A comprehensive discussion of kernel design via kernel composition is given by Duvenaud \cite{Duvenaud2014Thesis}.

A related idea to ours is the Group Automatic Relevance Determination kernel proposed by Xiong et al. \cite{10.1145/2964284.2967253}, where \gls{ard} parameters are tied across predefined feature groups.
This is close in spirit to the constructions considered here, since \gls{aprd} and \gls{acrd} can also be viewed as constrained \gls{ard} parametrisations.

Van der Wilk et al. \citep{VanDerWilk2017ConvGP} proposed a convolutional \gls{gp} model tailored to image-valued inputs.
Let $\vect{x} \in \mathbb{R}^D$ denote an image, let $\vect{x}^{[p]} \in \mathbb{R}^Q$ be the $p^{th}$ patch in the image and the image response function $f:\mathbb{R}^D\rightarrow \mathbb{R}$ be the weighted sum of local patch responses $h:\mathbb{R}^Q\rightarrow \mathbb{R}$.
Placing a \gls{gp} prior $h(\vect{z}) \sim \mathcal{GP}(0, k_h(\vect{z}, \vect{z}^\prime))$ on the patch response function gives
\begin{equation}\label{eq:wconvgp}
	\begin{aligned}
		f(\vect{x}) = \sum_{p=1}^{P} w_{p} h\left(\vect{x}^{[p]}\right) \Longrightarrow f(\vect{x}) \sim \mathcal{GP}\left(0, \sum_{p=1}^P \sum_{p^{\prime}=1}^P w_{p} w_{p^{\prime}}k_{h}\left(\vect{x}^{[p]}, \vect{x}^{\prime[p^\prime]}\right)\right).
	\end{aligned}
\end{equation}
Here, \(w_p\) is a patch-location weight that controls the contribution of a patch with index \(p\). 
For the present work, the important point is that the base kernel \(k_h\) is defined on image patches. 
This makes it possible to incorporate our proposed channel-wise relevance determination into convolutional \glspl{gp}.

\section{Kernel Constructions}\label{sec:kernel-construction}

\subsection{Automatic Patch Relevance Determination and Automatic Channel Relevance Determination}\label{sec:aprd-acrd}

If only the relevance of pixel positions is considered the input dimensions can be partitioned into disjoint sets where each set contains the colour channel values of a pixel.
For an image of shape $W\times H \times C$ there are $W \cdot H$ pixels.
The kernel is then constructed as the product
\begin{equation}\label{eq:pixelkernel}
	k(\vect{x}, \vect{x}^\prime) = \prod_{i=1}^{WH} k_i\left(\vect{x}^{[i]}, \vect{x}^{\prime[i]}\right) = \sigma^2\exp\left(-{\sum_{i=1}^{WH} \frac{\left\lVert \vect{x}^{[i]} - \vect{x}^{\prime[i]} \right\rVert^2}{2{\ell_i}^2}}\right)
\end{equation}
of isotropic \gls{rbf} kernels $k_i$.
The vector $\vect{x}^{[i]} \in \mathbb{R}^C$ contains the values of the pixel at location $i$.
The single hyperparameter $\sigma^2$ replaces the product of the $W\cdot H$ variance parameters from the isotropic \gls{rbf} kernels.
The total number of lengthscales is reduced to $L = W\cdot H$ which is the same number of lengthscales that a \gls{rbf}-\gls{ard} for single channel images would have.
By constraining the colour channel values of a pixel to share the same lengthscale, the model can no longer assign a separate lengthscale to an individual channel at this pixel.

Further reductions in the number of lengthscales are possible by partitioning the image into $P$ disjoint patches of shape $w\times h \times C$.
For each patch a separate isotropic \gls{rbf} kernel is defined and we obtain the full kernel as before via product construction.
Let $\vect{x}^{[p]} \in \mathbb{R}^{whC}$ be the $p^{th}$ patch.
Then the full kernel is defined as
\begin{equation}\label{eq:aprdkernel}
	k_{\mathrm{APRD}}(\vect{x}, \vect{x}^\prime) = \prod_{p=1}^{P} k_p\left(\vect{x}^{[p]}, \vect{x}^{\prime[p]}\right) = \sigma^2\exp\left(-{\sum_{p=1}^{P} \frac{\left\lVert \vect{x}^{[p]} - \vect{x}^{\prime[p]} \right\rVert^2}{2{\ell_{p}}^2}}\right).
\end{equation}
We refer to this as the \gls{rbf}-\gls{aprd} kernel.
This constraint reduces the ability of the model to adapt to isolated pixel-level variation, since all pixels within a patch share the same lengthscale.
The number of lengthscales decreases proportional to the patch size.
Assuming that $W$ and $H$ are divisible by $w$ and $h$, respectively, the number of patches is $P=(W\cdot H)\cdot (w \cdot h)^{-1}$.
For example: With a patch shape of $2\times 2$ and an image with shape $32\times 32 \times 3$ the number of lengthscales is $L=256$, a significant reduction from the $L=3072$ of a standard anisotropic \gls{rbf} kernel.
With a patch shape of $4\times 4$ the number of lengthscales is just $L=64$.

When only the relevance of colour channels is of interest, the kernel can be constructed as the product of isotropic \gls{rbf} kernels defined on the individual colour channels.
The \gls{rbf}-\gls{acrd} kernel is defined as
\begin{equation}\label{eq:acrdkernel}
	k_{\mathrm{ACRD}}(\vect{x}, \vect{x}^{\prime}) = \prod_{c=1}^{C} k_c\left(\vect{x}^{[c]}, \vect{x}^{\prime~[c]}\right) = \sigma^2\exp\left(-{\sum_{c=1}^{C} \frac{\left\lVert \vect{x}^{[c]} - \vect{x}^{\prime[c]} \right\rVert^2}{2{\ell_c}^2}}\right)
\end{equation}
where $c$ stands for the $c^{th}$ image channel.
All input dimensions in the same channel share the same lengthscale.
The total number of lengthscales is just equal to the number $C$ of colour channels.
Since \gls{aprd} and \gls{acrd} are constrained anisotropic \gls{rbf} kernels, their kernel evaluation has the same asymptotic cost as a standard anisotropic \gls{rbf} kernel, while using fewer independent lengthscale parameters.

\subsection{Channel-Additive Gaussian Processes}\label{sec:cadd-gps}

Let $\vect{x}$ denote an image with $C$ channels and let $\set{C}=\{1,\dots,C\}$ be the corresponding channel index set. 
For a subset $\set{S}\subseteq \set{C}$, we write $\vect{x}^\set{S}$ for the restriction of $\vect{x}$ to the channels indexed by $\set{S}$. 
A channel-additive model can then be defined by assigning a function $f_\mathcal{S}$ to each subset of channels and summing these components.

The channel-additive function is defined as
\begin{equation}\label{eq:caddfunc}
	f_{\mathrm{cadd}}(\vect{x}) = \sum_{m=1}^{C} \sum_{\substack{\set{S}\subseteq \set{C} \\ \lvert\set{S}\rvert=m}} f_\set{S}\left(\vect{x}^\set{S}\right).
\end{equation}
The terms with $\lvert\set{S}\rvert=1$ describe the contributions from individual channels, while the terms with $\lvert\set{S} \rvert > 1$ describe interactions between multiple channels. 

If independent \gls{gp} priors are placed on the component functions, then their sum is again a \gls{gp} with the channel-additive covariance kernel
\begin{equation}\label{eq:caddkernel}
	k_{\mathrm{cadd}}\left(\vect{x},\vect{x}^{\prime}\right) = \sum_{m=1}^{C} \sum_{\substack{\set{S}\subseteq \set{C} \\ \lvert\set{S}\rvert=m}} k_{\set{S}}\left(\vect{x}^{\set{S}},\vect{x}^{\prime\set{S}}\right).
\end{equation}
This construction has $2^C-1$ component kernels. 
For images where $C$ is large, this may become impractical.
One may restrict the set of channel subsets.

When standard \gls{rbf}-\gls{ard} kernels are selected for the component kernels $k_\mathcal{S}$, the total number of lengthscales is $12,288$ for images of shape $32\times 32 \times 3$.
This illustrates that a naive \gls{ard} parametrisation of all component kernels can quickly become impractical.
Instead using \gls{rbf}-\gls{acrd} kernels for all components is a suitable choice.
Each first order kernel has one lengthscale, each second order kernel has two lengthscales and the third order kernel has three lengthscales and so on.
The total number of lengthscales for an image of shape $32\times 32 \times 3$ is $12$.
Each of the seven kernels has an additional hyperparameter variance which brings the total number of hyperparameters to a moderate $19$. 

\subsection{Combining APRD, ACRD and Channel-Additivity}\label{sec:combinations}

Naturally, in many cases the relevance of pixel location and of colour channels are both important to determine.
We can treat positions and channels separately and combine \gls{aprd} and \gls{acrd}.
First we partition the image into $P$ disjoint patches with shape $w \times h \times C$ which gives a set of $P$ patch lengthscales.
A second set contains the $C$ lengthscales for the colour channels.
Let $\vect{x}^{[pc]}\in \mathbb{R}^{wh}$ be the values in the $c^{th}$ channel of the $p^{th}$ patch.
The combination kernel is defined as
\begin{equation}\label{eq:combinedkernel}
	k_{\mathrm{combined}}(\vect{x}, \vect{x}^{\prime}) = \sigma^2\exp\left(-\sum_{p=1}^{P}{\sum_{c=1}^{C} \frac{\left\lVert \vect{x}^{[pc]} - \vect{x}^{\prime[pc]} \right\rVert^2}{2{\ell_p}^2{\ell_c}^2}}\right).
\end{equation}
This corresponds to first scaling the inputs to the kernel channel-wise and then patch-wise or vice versa.
The number of lengthscales is $L=(W\cdot H)\cdot (w \cdot h)^{-1} + C$ which is the sum of the number of lengthscales that a \gls{rbf}-\gls{aprd} kernel and a \gls{rbf}-\gls{acrd}  kernel would have.

Incorporating \gls{aprd} into a channel-additive kernel $k_{\mathrm{cadd}}$ is similarly simple.
Let $s:\mathbb{R}^D \rightarrow \mathbb{R}^D$ be the transformation that rescales all entries in patch $p$ by ${\ell}^{-1}_p$.
A channel-additive \gls{aprd} kernel is obtained by first applying the scaling function $s$ to the inputs and then evaluating the channel-additive kernel:
\begin{equation}\label{eq:caddaprdkernel}
	k_{\mathrm{caddAPRD}}(\vect{x}, \vect{x}^\prime) = k_{\mathrm{cadd}}(s(\vect{x}), s(\vect{x}^\prime)).
\end{equation}
The parameters of function $s$ become hyperparameters of the kernel.
The total number of hyperparameters is the sum of the channel-additive kernel's number of hyperparameters and the number $P$ of patch lengthscales. 
By sharing the patch lengthscales across the component kernels, their covariance structures are tied through a common set of spatial relevance parameters. 
This imposes the assumption that spatial relevance is shared across the channel components.

\section{Experiments}\label{sec:experiments}

The implementation used for the experiments is based on GPflow \cite{Matthews2017GPflow, VanDerWilk2020Interdomain}.
Our implementation is available at \url{https://github.com/hrosse/structrd-caddgp}.
Sparse variational \gls{gp} models \cite{Hensman2013, Hensman2015} were used for the classification tasks.
For optimisation of the \gls{elbo} the Adam optimiser \cite{Kingma2015Adam} was used.

\subsection{MNIST}\label{sec:mnist}

\begin{figure}[h!] 
	\begin{center}
		\includegraphics[width=.99\textwidth]{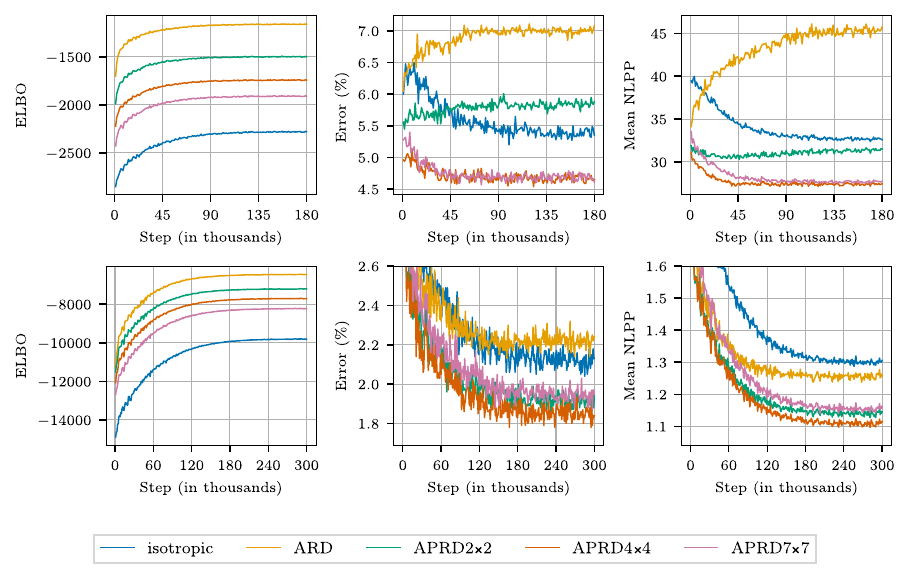} 
		\caption{Results for MNIST with 6,000 training points (top row) and 60,000 training points (bottom row) with isotropic \gls{rbf}, \gls{rbf}-\gls{ard} and \gls{rbf}-\gls{aprd} each with patches of shape $2\times 2$, $4\times 4$, and $7\times 7$. Left: evidence lower bound, middle: classification error, right: negative log predictive probability.} \label{fig:mnistcurves}
	\end{center}
\end{figure}

The MNIST dataset \cite{LeCun1998MNISTdataset} consists of images of handwritten digits split into 60,000 training points and 10,000 test points.
Each image is a grayscale image of size $28 \times 28$ pixels.
\gls{gp} models with five different kernels were trained: Isotropic \gls{rbf} kernel, \gls{rbf}-\gls{ard} and \gls{rbf}-\gls{aprd} with patch shapes $2\times 2$, $4\times 4$ and $7\times 7$ respectively.
In the first set of experiments the models were trained on a subset of the training data consisting of 6,000 randomly sampled points.
In the second set of experiments the complete training data was used.
Separate independent latent \glspl{gp} with 750 separate inducing points and separate kernels were used for each class and mapped to a SoftMax likelihood.
The initialisation was identical for each run.
An initial learning rate of 0.01 was selected and multiplied by 0.9 every 4,000 steps during optimisation with the smaller training data and multiplied by 0.9 every 5,000 steps with the full set of training points.
Minibatches of size 500 were used in each run.
Table \ref{tab:mnisttable} shows the final results. 
Figure \ref{fig:mnistcurves} shows \gls{elbo}, classification error and \gls{nlpp} with respect to the training steps.

\begin{table}
	\begin{center}
		\begin{tabular}{lccccc}\toprule
			& & \multicolumn{2}{c}{MNIST 6,000} & \multicolumn{2}{c}{MNIST 60,000}
			\\\cmidrule(lr){3-4}\cmidrule(lr){5-6}
			model & \# lengthscales & error (\%) & \gls{nlpp} ($\times 10^{2}$)   &   error (\%) & \gls{nlpp} ($\times 10^{2}$)\\ \midrule
			isotropic& $1$ & $5.35$ & $32.58$ & $2.13$ & $1.31$  \\
			\gls{ard}& $784$ & $7.08$ & $45.76$ & $2.24$ & $1.26$  \\
			\gls{aprd}(2$\times$2)& $196$ & $5.88$ & $31.49$ & $1.94$ & $1.15 $ \\
			\gls{aprd}(4$\times$4)& $49$ & $4.65$ & $27.46$ & $1.84$ & $1.12$ \\
			\gls{aprd}(7$\times$7)& $16$ & $4.61$ & $27.73$ & $1.95$ & $1.16$  \\ 
			\bottomrule
		\end{tabular}
	\end{center}
	\caption{Test results for MNIST. The number of lengthscales is reported per latent \gls{gp}.} \label{tab:mnisttable}
\end{table}
For the experiment with $6,000$ training points, full \gls{ard} achieves the highest \gls{elbo}.
At the same time its classification error and \gls{nlpp} deteriorate during training.
This behaviour is consistent with pixel-wise overfitting in a small-data regime.
The isotropic \gls{rbf} kernel performs better than full \gls{ard} in terms of predictive metrics.
The \gls{aprd} variants improve the \gls{nlpp}, with the coarser $4\times 4$ and $7\times 7$ patch structures reducing classification error and showing the most stable \gls{nlpp} curves.

With the full MNIST training set, the behaviour of full \gls{ard} changes.
The \gls{nlpp} no longer deteriorates and becomes lower than that of the isotropic \gls{rbf} kernel, suggesting that the larger amount of data partly mitigates the instability previously observed. 
All \gls{aprd} variants still achieve lower \gls{nlpp} than full \gls{ard}.
In this setting, finer patch structures become more competitive.
The $4\times 4$ variant has the best overall predictive performance in this experiment.

\subsection{Planes in Satellite Imagery}\label{sec:planes}

\begin{figure}[h!] 
	\begin{center}
		\includegraphics[width=.99\textwidth]{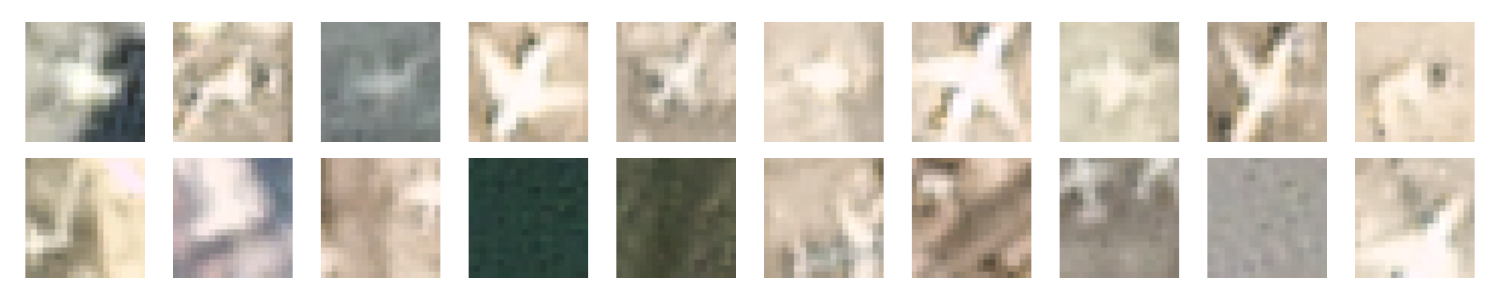} 
		\caption{Sample from the Planes in Satellite Imagery dataset. Top row: class 1, bottom row: class 0.} \label{fig:planes_examples}
	\end{center}
\end{figure}

The Planes in Satellite Imagery dataset \cite{robert_hammell_2023} consists of 32,000 RGB satellite images of size $20 \times 20 \times 3$.
The images are divided into two classes:
Class 1 are 8,000 images that contain an airplane in the centre of the image.
Class 0 consists of 24,000 images that either contain no airplane or contain airplanes that are not centred. 
The negative examples are divided into three groups: randomly cropped land features, partial airplanes, and bright features that are visually similar to airplanes.
The images were acquired under varying atmospheric conditions.
A small sample from the dataset is shown in figure \ref{fig:planes_examples}.
The dataset was split into 75\% training data (24,000 points) and 25\% test data (8,000 points).
We perform two sets of experiments on this dataset. 
The first compares the proposed kernels defined on full images, while the second investigates convolutional \gls{gp} models and their combination with the proposed structured kernels.

\subsubsection{Experiment 1}\label{sec:planes-exp1}

\begin{figure}[h!] 
	\begin{center}
		\includegraphics[width=.99\textwidth]{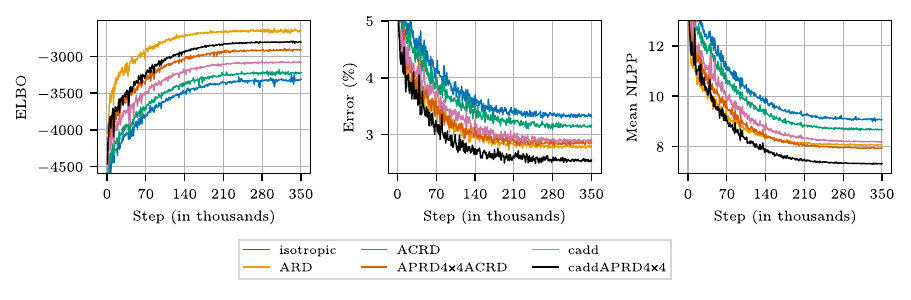} 
		\caption{Results for experiment 1 on Planes in Satellite Imagery dataset. Left: evidence lower bound, middle: classification error, right: negative log predictive probability.} \label{fig:planescurve1}
	\end{center}
\end{figure}

Six \gls{gp} models with different kernels were trained: Isotropic \gls{rbf} kernel, \gls{rbf}-\gls{ard}, \gls{rbf}-ACRD, combination of \gls{rbf}-\gls{acrd} and \gls{aprd} with patch shape $4\times 4$, a channel-additive kernel with \gls{rbf}-\gls{acrd} base kernels and the combination of a channel-additive kernel with \gls{rbf}-\gls{acrd} base kernels and \gls{aprd} with patch shape $4\times 4$.
The number of inducing points was 750 and the variational parameters were identically initialised across all runs.
For binary classification a Bernoulli likelihood was chosen.
An initial learning rate of 0.01 was selected and multiplied by 0.9 every 5,000 steps.
The minibatch size was 500 for each run.
Final test results are shown in Table \ref{tab:planestable}. 
Figure \ref{fig:planescurve1} shows \gls{elbo}, classification error and \gls{nlpp} with respect to the training steps.

\begin{table}
	\begin{center}
		\begin{tabular}{lccc}\toprule
			model & \# kernel hyperparameters  & error (\%) & \gls{nlpp} ($\times 10^{2}$)\\ \midrule
			isotropic& $2$ & $3.31$ & $9.07$   \\
			\gls{ard} & $1201$ & $2.8$ & $8.08$   \\
			\gls{acrd}& $4$ & $3.16$ & $8.67$  \\
			\gls{aprd}(4$\times$4) + \gls{acrd}& $29$ & $2.86$ & $7.94$  \\
			channel-additive& $19$ & $2.89$ & $8.19$   \\ 
			channel-additive \gls{aprd} (4$\times$4)& $44$ & $2.55$ & $7.32$   \\ 
			\bottomrule
		\end{tabular}
	\end{center}
	\caption{Test results for experiment 1 on Planes in Satellite Imagery dataset.} \label{tab:planestable}
\end{table}
The results show that incorporating channel-additive structure is beneficial. 
The \gls{rbf}-\gls{acrd} kernel improves both classification error and \gls{nlpp} compared to the isotropic \gls{rbf} kernel.
The channel-additive kernel with \gls{rbf}-\gls{acrd} base kernels improves both predictive metrics which suggests that contributions from individual channels and channel subsets can explain important parts of the function.
The full \gls{rbf}-\gls{ard} kernel achieves a similar result to the channel-additive kernel.
The combination of \gls{aprd} with \gls{acrd} slightly improves the \gls{nlpp} relative to full \gls{ard}.
The channel-additive \gls{aprd} kernel yields the strongest overall performance in this experiment, with a notable reduction in both classification error and \gls{nlpp}.

\subsubsection{Experiment 2}\label{sec:planes-exp2}

The first two convolutional \gls{gp} models are defined as follows: One with isotropic \gls{rbf} base kernel and one with \gls{rbf}-\gls{acrd} base kernel.
The other two \gls{gp} models are composed from in the prior independent latent processes: 
One model is defined as the sum of a convolutional \gls{gp} with an isotropic base kernel and a \gls{gp} with \gls{rbf}-\gls{ard} kernel, the other model is defined as the sum of a convolutional \gls{gp} with \gls{acrd} base kernel and a \gls{gp} with channel-additive \gls{aprd} kernel, i.e. $f(\vect{x}) = f_{conv}(\vect{x}) + f_{ARD}(\vect{x})$ and $f(\vect{x}) = f_{convACRD}(\vect{x}) + f_{caddAPRD}(\vect{x})$.
For the convolutional kernels a patch shape of $6\times 6$ is chosen, for \gls{aprd} the patch shape is $4\times 4$.

\begin{figure}[h!] 
	\begin{center}
		\includegraphics[width=.99\textwidth]{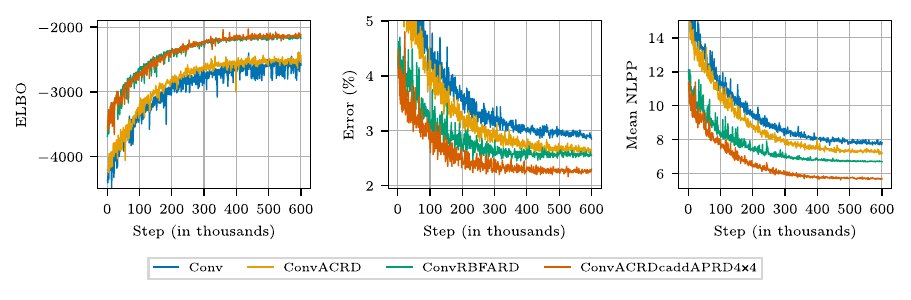} 
		\caption{Results for experiment 2 on Planes in Satellite Imagery dataset. Left: evidence lower bound, middle: classification error, right: negative log predictive probability.} \label{fig:planescurve2}
	\end{center}
\end{figure}

Following the inter-domain approach proposed by van der Wilk et al. \citep{VanDerWilk2017ConvGP}, the inducing points of the latent convolutional processes are placed in the space of patches.
For the models consisting of two latent \glspl{gp}, the corresponding posteriors are additively mixed prior to the likelihood under a mean-field approximation.
The models were initialised with 750 separate inducing points per latent process.
The minibatch size was 240 for each run.
An initial learning rate of 0.01 was multiplied by 0.93 every 7,000 steps.
Table \ref{tab:planestable-exp2} shows the final test results, figure \ref{fig:planescurve2} shows \gls{elbo}, classification error and \gls{nlpp} with respect to training steps.

\begin{table}
	\begin{center}
		\begin{tabular}{lcc}\toprule
			model & error (\%) & \gls{nlpp} ($\times 10^{2}$)\\ \midrule
			Convolutional & $2.85$ & $7.82$   \\
			Convolutional-\gls{acrd}& $2.61$ & $7.19$  \\
			Convolutional + \gls{rbf}-\gls{ard}& $2.55$ & $6.69$   \\ 
			Convolutional-\gls{acrd} + channel-additive \gls{aprd} (4x4)& $2.3$ & $5.69$   \\ 
			\bottomrule
		\end{tabular}
	\end{center}
	\caption{Test results for experiment 2 on Planes in Satellite Imagery dataset.} \label{tab:planestable-exp2}
\end{table}

The convolutional \gls{gp} with an \gls{rbf}-\gls{acrd} base kernel improves both classification error and \gls{nlpp} compared to the convolutional \gls{gp} with an isotropic \gls{rbf} base kernel. 
This suggests that determining channel-specific relevance is also beneficial when used on the level of local patch-responses. 
Adding a non-convolutional component further improves predictive performance. 
The strongest result is obtained by combining a convolutional \gls{gp} with an \gls{rbf}-\gls{acrd} base kernel and a channel-additive \gls{aprd} component. 
This model achieves the lowest classification error and, more prominently, the lowest \gls{nlpp} among all models considered. 
The result suggests that the inductive bias provided by convolutional structure and the parameter-efficient, channel-additive non-local structure capture complementary aspects of the latent function.

\section{Discussion}\label{sec:discussion}

The MNIST experiments illustrate that \gls{aprd} is a beneficial compromise between the isotropic \gls{rbf} kernel and full \gls{ard}. 
By tying lengthscales across spatial patches \gls{aprd} preserves spatially varying relevance while reducing the ability of the model to adapt to isolated pixel-level variation. 
Across both MNIST data regimes, \gls{aprd} improves \gls{nlpp} compared to full \gls{ard}. 
Larger patches are more stable with fewer training points, while finer patches become more competitive when the full training set is used.
Candidate patch sizes can be compared using held-out \gls{nlpp}, which evaluates the quality of probabilistic predictions and can reveal instability that is not visible from the \gls{elbo} alone.

Applying lengthscale sharing only at the level of colour channels already improved over the isotropic \gls{rbf} kernel in the Planes in Satellite Imagery experiment.
The \gls{rbf}-\gls{acrd} kernel can mitigate strong intra-class variations in a colour channel by a larger channel-specific lengthscale.

One possible explanation for the improvement of the channel-additive kernels is that they can preserve partial similarities across channel subsets.
For example, two images may contain a similarly shaped bright airplane, while differing strongly in one colour channel due to background or variation in atmospheric conditions.
\gls{rbf} kernels over all channels depend on a single Euclidean distance over the full input, so differences across many dimensions are accumulated in one term and can strongly reduce the kernel value.
The channel-additive kernel in contrast can still assign high similarity through the components that are restricted to a subset of channels.
More generally, the channel-additive kernel combines two aspects: 
The highest-order \gls{rbf} component on the full input only encodes a smoothness assumption.
The lower-order components provide structured lower-dimensional explanations by modelling separate functions on meaningful subspaces of the input.
The marginal likelihood provides a mechanism to adjust the relative contribution of the component functions through the variance parameters of the component kernels.
This is achieved with relatively few additional hyperparameters by sharing patch relevance parameters across additive components instead of assigning separate spatial lengthscales to every component.

The convolutional \gls{gp} experiments show that the proposed hyperparameter-sharing constructions can also be incorporated into established image-specific \gls{gp} priors. 
In a convolutional \gls{gp}, the base kernel is defined on local image patches.
The structural assumptions at the image level can be applied locally at the patch level.
Since the patches retain the channel structure of the original image, channel-wise lengthscale sharing can be applied at the level of local patch responses by choosing an \gls{rbf}-\gls{acrd} base kernel. 
The improvement of the convolutional \gls{gp} with an \gls{rbf}-\gls{acrd} base kernel over the convolutional \gls{gp} with an isotropic \gls{rbf} base kernel suggests that channel-specific lengthscales are useful not only for kernels on full images, but also for local patch-response models. 
This is consistent with the interpretation that strong channel-wise variation present in the full images also appears within local patches.

The underlying principle is not specific to images. 
Similar structures may be found in other domains.
The same idea can be applied when the input consists of repeated or structured components with comparable internal variables.

\section{Conclusion}\label{sec:conclusion}

We introduced structured relevance determination kernels for \glspl{gp} on image-valued inputs. \gls{aprd} and \gls{acrd} constrain the \gls{ard} parametrisation of the \gls{rbf} kernel by sharing lengthscales across spatial patches and image channels, respectively. These constructions reduce the number of independently learned lengthscales while preserving the ability to model spatially and channel-wise varying relevance. We further introduced channel-additive \glspl{gp} that decompose the latent function into components defined on channel subsets and combined this construction with patch-wise relevance determination. We also showed how channel-wise relevance determination can be incorporated into convolutional \glspl{gp} by using an \gls{rbf}-\gls{acrd} base kernel on image patches.

The experiments show that structured lengthscale sharing can improve predictive performance compared to both the isotropic \gls{rbf} kernel and full \gls{ard}. On MNIST, \gls{aprd} improved \gls{nlpp} in both the small-data and full-data regimes, with the preferred patch size depending on the amount of available data. On the Planes in Satellite Imagery dataset, channel-additive \gls{aprd} achieved the strongest result among the full-image kernels. The convolutional \gls{gp} experiments further show that \gls{acrd}  can improve a convolutional \gls{gp} when used as the patch-level base kernel. The strongest overall result was obtained by combining a convolutional \gls{gp} with an \gls{rbf}-\gls{acrd} base kernel and a channel-additive \gls{aprd} component, suggesting that local convolutional structure and parameter-efficient non-local channel-subset structure capture complementary aspects of the latent function.

These results suggest that the practical effectiveness of inductive biases and non-local generalisation mechanisms in \glspl{gp} depends not only on whether the chosen structure is suitable, but also on how parameter-efficiently it is implemented. Future work could investigate more flexible or learned grouping structures, apply the proposed ideas to other structured high-dimensional inputs, and further combine hyperparameter sharing with stronger problem-specific inductive biases.

\bibliographystyle{plainnat}
\bibliography{references}

\end{document}